\documentclass[conference,fleqn]{IEEEtran}
\IEEEoverridecommandlockouts

\usepackage{amssymb}
\usepackage{amsmath,amssymb,amsfonts} 
\usepackage{graphicx}
\usepackage{textcomp}
\usepackage{cite}
\usepackage[caption=false,font=footnotesize]{subfig}

\def\BibTeX{{\rm B\kern-.05em{\sc i\kern-.025em b}\kern-.08em
    T\kern-.1667em\lower.7ex\hbox{E}\kern-.125emX}} 

\usepackage{amsmath}
\usepackage{bm}
\usepackage[hidelinks]{hyperref}
\usepackage{orcidlink}

\newcommand\copyrighttext{%
  \footnotesize \textcopyright 2021 IEEE. Personal use of this material is permitted. Permission from IEEE must be obtained for all other uses, in any current or future media, including reprinting/republishing this material for advertising or promotional purposes, creating new collective works, for resale or redistribution to servers or lists, or reuse of any copyrighted component of this work in other works. IEEE Copyright policy can be found on \textit{\href{https://www.ieee.org/publications/rights/copyright-policy.html}{https://www.ieee.org/publications/rights/copyright-policy.html}}
  }
\newcommand\copyrightnotice{%
\begin{tikzpicture}[remember picture,overlay]
\node[anchor=south,yshift=10pt] at (current page.south) {\fbox{\parbox{\dimexpr\textwidth-\fboxsep-\fboxrule\relax}{\copyrighttext}}};
\end{tikzpicture}%
}

\begin{document}

\title{Reinforcement Learning Algorithms: An Overview and Classification}

\author{
\IEEEauthorblockN
{
    Fadi AlMahamid~\orcidlink{0000-0002-6907-7626},~\textit{Senior Member, IEEE}, and Katarina Grolinger~\orcidlink{0000-0003-0062-8212},~\textit{Member, IEEE}
}
\IEEEauthorblockA
{
    \textit{Department of Electrical and Computer Engineering}\\
    \textit{Western University}\\
    London, Ontario, Canada\\
    Email: \{falmaham, kgroling\}@uwo.ca
}
}
\maketitle
\copyrightnotice

\vspace{-10pt}
\begin{abstract}
The desire to make applications and machines more intelligent and the aspiration to enable their operation without human interaction have been driving innovations in neural networks, deep learning, and other machine learning techniques. Although reinforcement learning has been primarily used in video games, recent advancements and the development of diverse and powerful reinforcement algorithms have enabled the reinforcement learning community to move from playing video games to solving complex real-life problems in autonomous systems such as self-driving cars, delivery drones, and automated robotics. Understanding the environment of an application and the algorithms' limitations plays a vital role in selecting the appropriate reinforcement learning algorithm that successfully solves the problem on hand in an efficient manner. Consequently, in this study, we identify three main environment types and classify reinforcement learning algorithms according to those environment types. Moreover, within each category, we identify relationships between algorithms. The overview of each algorithm provides insight into the algorithms' foundations and reviews similarities and differences among algorithms. This study provides a perspective on the field and helps practitioners and researchers to select the appropriate algorithm for their use case.
\end{abstract}
\IEEEpeerreviewmaketitle

\section{Introduction} \label{sec:introduction}
Reinforcement Learning (RL) is one of the three machine learning paradigms besides supervised learning and unsupervised learning. It uses agents acting as human experts in a domain to take actions. RL does not require data with labels; instead, it learns from experiences by interacting with the environment, observing, and responding to results.

\begin{figure}[!t]
    \centering
    \includegraphics[width=.7\linewidth]{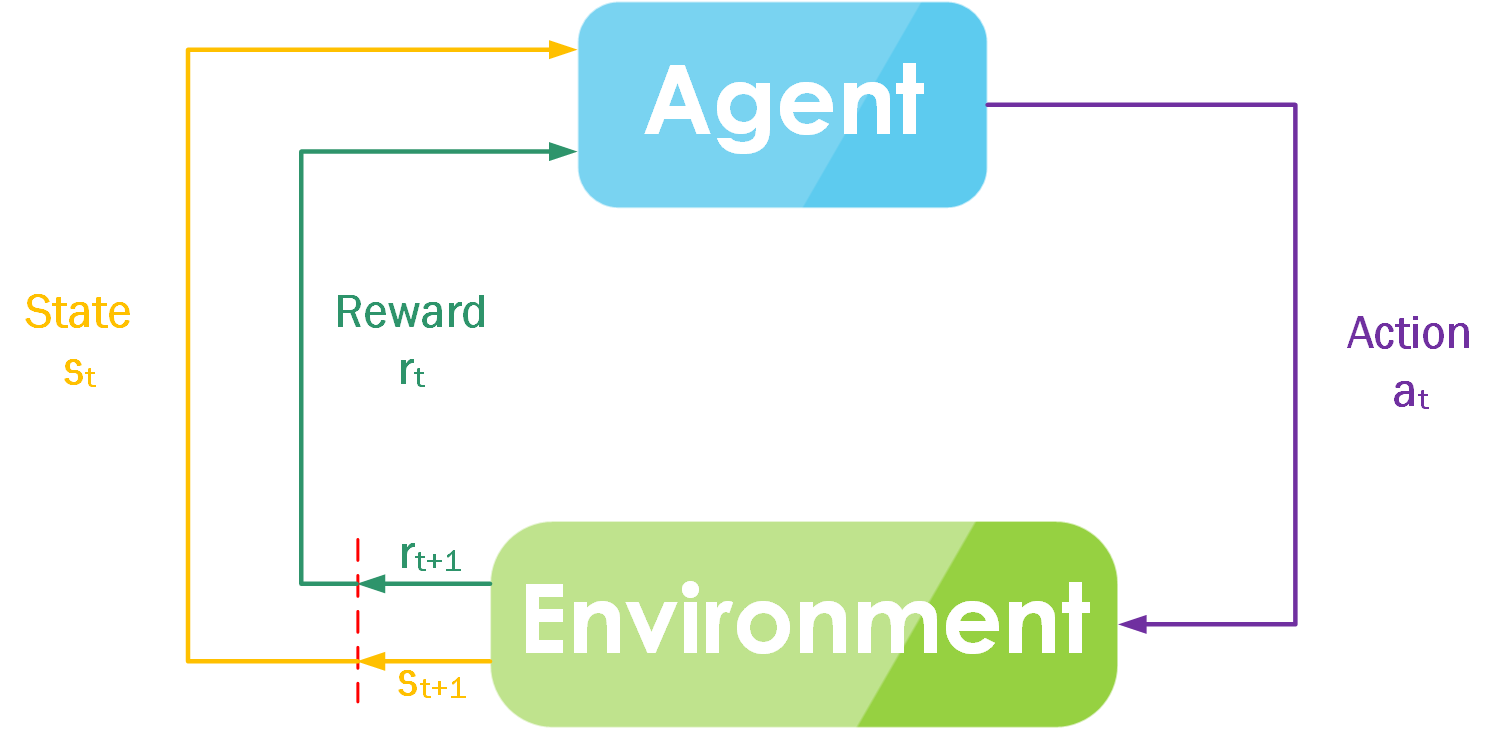}
    \caption{Markov Decision Process}
    \label{fig:mdp-arch}
\end{figure}

RL can be expressed with Markov Decision Process (MDP) as shown in Figure \ref{fig:mdp-arch}. Each environment is represented with a state that reflects what is happening in the environment. The RL agent takes actions in the environment, that causes a change in the environment's current state generating a new state and receives a reward based on the results. The agent receives a positive reward for good actions and a negative reward for bad actions, which helps the agent evaluate the performed action in a given state and learn from experiences.

Video games have been one of the most popular RL applications, and RL algorithms have been mainly tested and evaluated on video games. However, RL has other applications and can be used in different domains such as self-driving cars, natural language processing (NLP), autonomous robotics, delivery drones, and many others. Furthermore, there are many diverse RL algorithms with different variations. Therefore, it is imperative to understand the differences between RL algorithms, select the appropriate algorithm suitable for the environment type and the task on hand.

The most widely used algorithm is Deep Q-Network (DQN) with its variations because of its simplicity and efficiency. Nevertheless, DQN is suitable only for environments with discrete actions. For example, in autonomous UAV navigation (self-flying drones), many papers tend to simplify the environment to enable the use of DQN \cite{okuyama2018autonomous, chishti2018self}. However, in complex real-life environments, DQN would not be suitable if the environment is dynamic or the required actions are continuous. Therefore, to assist in matching the RL algorithm with the task, the classification of RL algorithms based on the environment type is needed.

Consequently, this study provides an overview of different RL algorithms, classifies them based on the environment type, and explains their primary principles and characteristics. Additionally, relationships among different RL algorithms are also identified and described. The paper provides a perspective on the domain and helps researchers and practitioners to select appropriate algorithms for their use cases. Moreover, it provides options for selecting a suitable algorithm for the environment, rather than attempting to simplify the environment for the algorithm \cite{okuyama2018autonomous, chishti2018self}.  

The remainder of the paper is organized as follows: Section \ref{sec:background} introduces RL and discusses RL's main principles. Section \ref{sec:rl-algorithms} classifies RL algorithm and provides their overview. Finally, Section \ref{sec:conclusion} concludes the paper.

\section{Background} \label{sec:background}
This section first introduces reinforcement learning. Next, concepts of policy and value functions are described, and finally, experience replay is explained as a commonly used technique in different RL algorithms.

\subsection{Reinforcement Learning}
The RL agent learns from taking actions in the environment, which causes a change in the environment's current state and generates a reward based on the action taken as expressed in the Markov Decision Process (MDP). We define the probability of the transition to state $\mathbf{s^{\prime}}$ with reward $\bm{r}$ from taking action $\bm{a}$ in state $\bm{s}$ at time $\bm{t}$, for all $s^{\prime} \in S, \; s \in S, \; r \in R, \; a \in A(s)$, as:

\begin{equation}
    P(s^{\prime},r|s,a) = Pr\{S_t = s^{\prime}, R_t=r| S_{t-1}=s, A_{t-1}=a\}    
\end{equation}

The agent receives rewards for performing actions and uses them to measure the action's success or failure. The Reward $R$ can be expressed in different forms, as a function of the action $R(a)$, or as a function of action-state pairs $R(a,s)$.

The agent's objective is to maximize the expected summation of the discounted rewards, which drives the agent to take the selected actions. The reward is granted by adding all the rewards generated from executing an episode. The \textit{episode} (trajectory) represents a finite number of actions and ends when the agent achieves a final state, for example, when a collision occurs in a simulated navigation environment. However, in some cases, the actions can be continuous and cannot be broken into episodes. The discounted reward, as shown in equation \ref{equ:sum-disc-reward} uses a multiplier $\bm{\gamma}$ to the power $\bm{k}$, where $\bm{\gamma \in [0,1]}$. The value of $k$ increases by one at each time step to emphasize the current reward and to reduce the impact of the future rewards, hence the term discounted reward.

\begin{equation}
    \label{equ:sum-disc-reward}
    G_t = E \left[\sum_{k=0}^{\infty} \gamma^{k}  R_{t+k+1}\right]
\end{equation}

Emphasizing the current action’s immediate reward and reducing the impact of future actions’ rewards help the expected summation of discounted rewards to converge.

\subsection{Policy and Value Function}
The agent’s behavior is defined by following a policy $\bm{\pi}$, where the policy $\bm{\pi}$ defines the probability of taking action $\bm{a}$, given a state $\bm{s}$, which is denoted as $\bm{\pi(a|s)}$. Once the agent takes an action, the agent uses a value function to evaluate the action. The agent either uses: 1) a \textit{state-value function} to estimate how good for the agent to be in state $\bm{s}$, or 2) a \textit{action-value function} to measure how good it is for the agent to perform an action $\bm{a}$ in a given state $\bm{s}$. The action-value function is defined in terms of the expected summation of the discounted rewards and represents the target Q-value:
\begin{equation}
    \label{equ:action-value-func}
    Q_\pi(s,a) = E_\pi \left[\sum_{k=0}^{\infty} \gamma^{k}  R_{t+k+1} \; | \; S_t = s, A_t = a \right]
\end{equation}

The agent performs the action with the highest Q-value, which might not be the optimal Q-value. Finding the optimal Q-value requires selecting the best actions that maximize the expected summation of discounted rewards under the optimal policy $\bm{\pi}$. The optimal Q-value $\bm{Q_*(s,a)}$ as described in equation \ref{equ:optimal-policy} must satisfy the \textit{Bellman optimality equation}, which is equal to the expected reward $\bm{R_{t+1}}$, plus the maximum expected discounted return that can be achieved for any possible next state-action pairs $\bm{(s^\prime,a^\prime)}$.

\begin{equation}
    \label{equ:optimal-policy}
    Q_*(s,a) = \underset{\pi}{max} \; Q(s,a)
\end{equation}

\vspace{-10pt}
\begin{equation}
    \label{equ:bellman-optimality}
    Q_*(s,a) = E \left[ R_{t+1}  + \gamma \; \underset{a^\prime}{max} \; Q_*(s^\prime,a^\prime) \right]  
\end{equation}

This optimal Q-value $\bm{Q_*(s,a)}$ is used to train the neural network. The Q-value $\bm{Q(s,a)}$ predicted by the network is subtracted from the optimal Q-value $\bm{Q_*(s,a)}$ estimated using the Bellman equation and backpropagated through the network. The loss function is defined as follows: 

\begin{equation}
    \label{equ:loss-func}
    \overset{Target}{\overbrace{E \left[ R_{t+1}  + \gamma \; \underset{a^\prime}{max} \; Q_*(s^\prime,a^\prime) \right]}} \; - \;  \overset{Predicted}{\overbrace{E \left[\sum_{k=0}^{\infty} \gamma^{k}  R_{t+k+1}\right]}}
\end{equation}

\subsection{Experience Replay}
In RL, an experience $\bm{e}$ can be described as the knowledge produced from the agent performing an action $\bm{a}$ in a state $\bm{s}$ causing a new state $\bm{s^\prime}$ and a generated reward $\bm{r}$. The experience can be expressed as a tuple $\bm{e(s,a,s^\prime,r)}$. Lin \cite{lin1992self} proposed a technique called \textit{Experience Replay}, where experiences are stored in a replay memory $\bm{D}$ and used to train the agent. Since experiences are stored in the memory, and some experiences might be of a high importance, they can repeatedly be reused to train the agent what improves convergence.

Although experience replay should help the agent theoretically learn from previous important experiences, it entails sampling experiences uniformly from the replay memory $\bm{D}$ regardless of their significance. Schaul \textit{et al.}  \cite{schaul2015prioritized} suggested the use of \textit{Prioritized Experience Replay}, which aims to prioritize experiences using Temporal Difference error (TD-error) and replay more frequently experiences that have lower TD-error.

\section{Reinforcement Learning Algorithms Classification} \label{sec:rl-algorithms}
\begin{figure*}[!t]
    \centering
    \includegraphics[width=.7\linewidth]{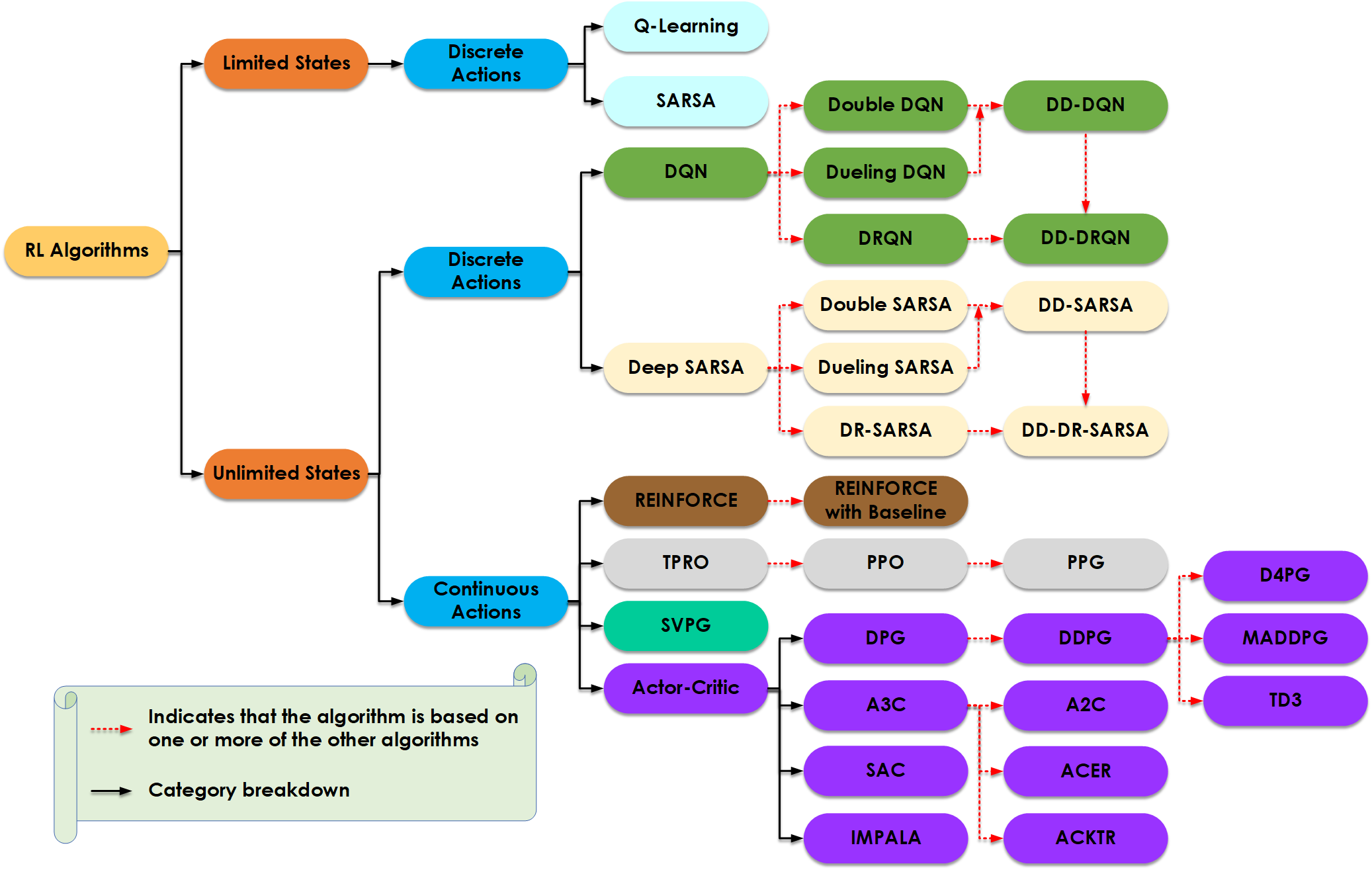}
    \caption{Reinforcement Algorithms classification based on the environment type}
    \label{fig:RL-Algorthims}
\end{figure*}

While most reinforcement learning algorithms use deep neural networks, different algorithms are suited for different environment types. We classify RL algorithms according to the number of the states and action types available in the environment into three main categories: 1) a limited number of states and discrete actions, 2) an unlimited number of states and discrete actions, and 3) an unlimited number of states and continuous actions. The three categories, together with algorithms belonging to those categories, are shown in Figure \ref{fig:RL-Algorthims} and discussed in the following subsections.

\subsection{Environments with Limited States and Discrete Actions} \label{LimitedStatesDiscreteActions}
The environments with discrete actions and limited states are relatively simple environments where the agent can select from pre-defined actions and be in pre-defined known states. For example, when an agent is playing a tic-tac-toe game, the nine boxes represent the states, and the agent can choose from two actions: X or O, and update the available states.

Q-Learning \cite{watkins1992q} algorithm is commonly used to solve problems in such environments. This algorithm finds the optimal policy in a Markov Decision Process (MDP) by maintaining a Q-Table table with all possible states and actions and iteratively updating the Q-values for each state-action pair using the Bellman equation until the Q-function converges to the optimal Q-value.

State–Action–Reward–State–Action (SARSA) \cite{rummery1994line} is another algorithm from this category: it is similar to Q-learning except it updates the current $\bm{Q(s,a)}$ value in a different way. In Q-learning, in order to update the current $\bm{Q(s,a)}$ value, we need to compute the next state-action $\bm{Q(s^\prime,a^\prime)}$ value, and since the next action is unknown, then Q-learning takes a greedy action to maximize the reward \cite{zhao2016deep}. In contrast, when SARSA updates the current state-action $\bm{Q(s,a)}$ value, it performs the next action $\bm{a^\prime}$ \cite{zhao2016deep}.

\subsection{Environments with Unlimited States and Discrete Actions}
In some environments, such as playing a complex game, the states can be limitless; however, the agent's choice is limited to a finite set of actions. In such environments, the agent mainly consists of a Deep Neural Network (DNN),  usually a Convolutional Neural Network (CNN), responsible for processing and extracting features from the state of the environment and outputting the available actions. Different algorithms can be used with this environment type, such as Deep Q-Networks (DQN), Deep SARA, and their variants.

\vspace{6 pt}
\subsubsection{\textbf{Deep Q-Networks (DQN)}}
\begin{figure}[!t]
    \centering
    \includegraphics[width=1\linewidth]{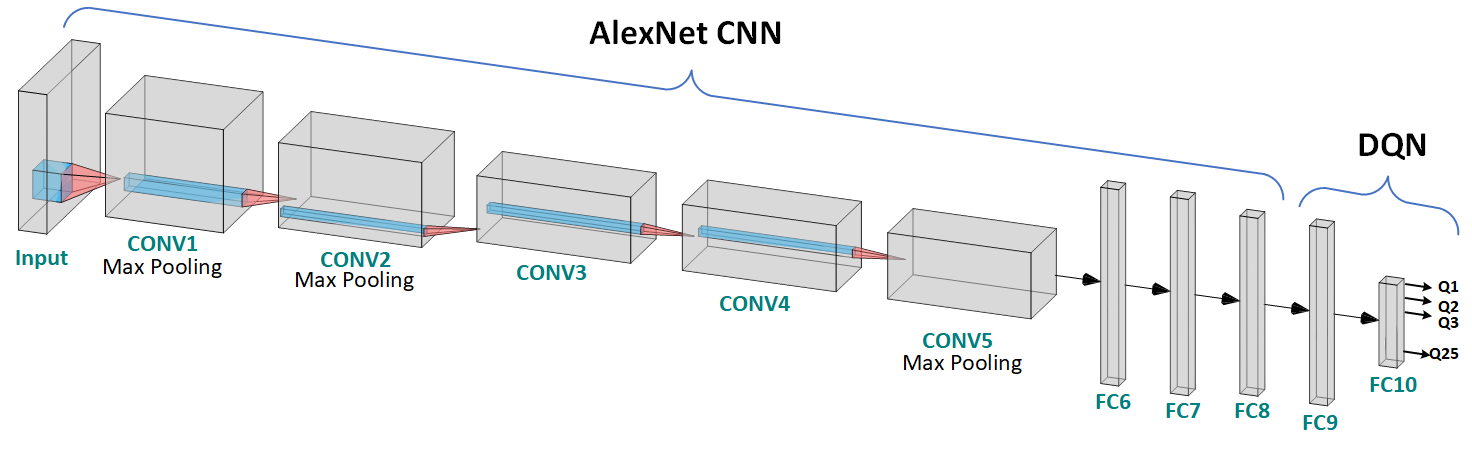}
    \caption{DQN using AlexNet CNN}
    \label{fig:dqn}
\end{figure}

\hfill \break
Deep Q-Learning, also referred to as Deep Q-Networks (DQN), is considered the main algorithm used in environments with unlimited states and discrete actions, and it inspires other algorithms used for a similar purpose. DQN usually combines convolutional and pooling layers, followed by fully connected layers that produce Q-values corresponding to the number of actions. Figure \ref{fig:dqn} \cite{anwar2020autonomous} shows AlexNet CNN followed by two fully connected layers to produce Q-value. The current scene from the environment represents the environment's current state; once it is passed to the network, it produces Q-value representing the best action to take. The agent acts and then captures the changes in the environment's current state and the reward generated from the action.

A significant drawback of the DQN algorithm is overestimating the action-value (Q-value), where the agent tends to choose a non-optimal action because it has the highest Q-value \cite{VanHasselt2010}.
\vspace{3 pt}
\paragraph{Double and Dueling DQN (DD-DQN)}
\hfill \break
Double DQN uses two networks to solve this overestimation problem in DQN. The first network, called the Policy Network, optimizes the Q-value, and the second network, the Target Network, is a replica of the policy network, and it is used to produce the estimated Q-value \cite{VanHasselt2016}. The target network parameters are updated after a certain number of time steps by copying the policy network parameters rather than using the backpropagation. 

\begin{figure}[!t]
    \centering
    \includegraphics[width=1\linewidth]{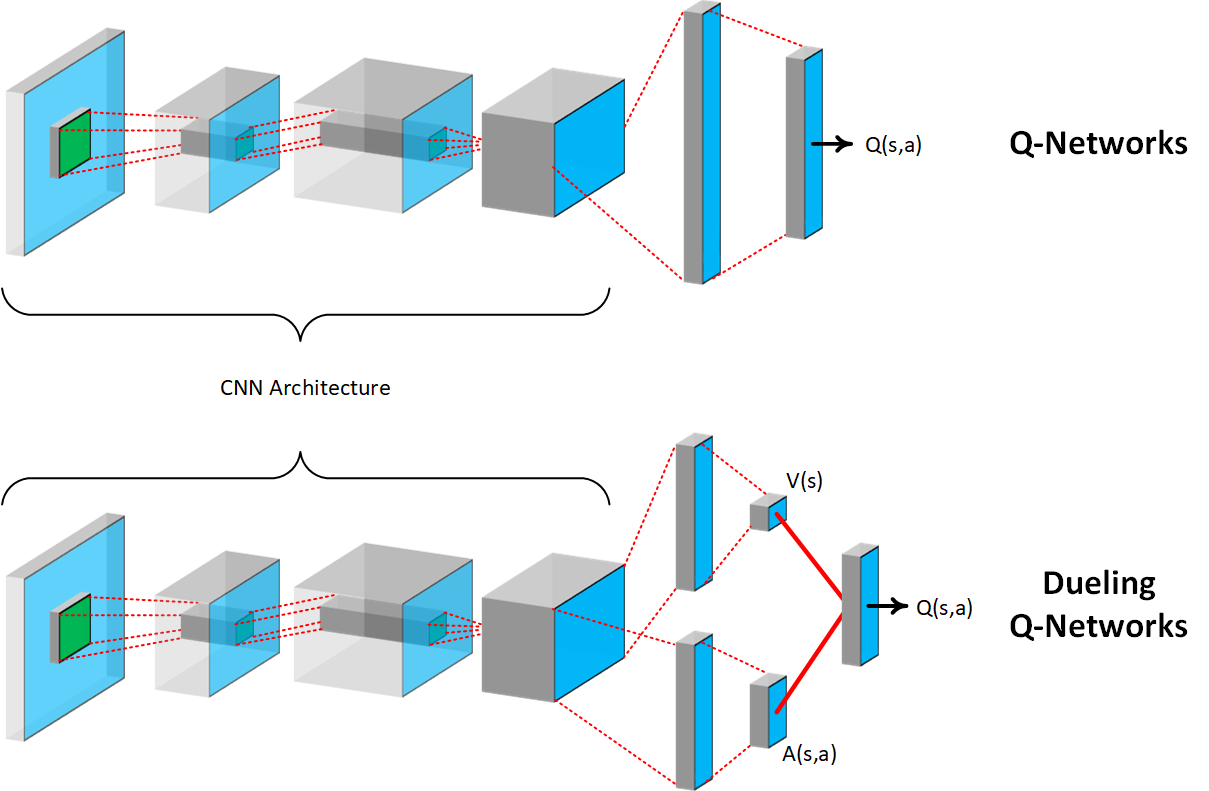}
    \caption{DQN vs. Dueling DQN}
    \label{fig:Dueling-DQN}
\end{figure}

Another improvement on DQN is Dueling DQN illustrated in Figure \ref{fig:Dueling-DQN} \cite{Wang2016}. Dueling DQN tries to define a better way to evaluate the Q-value by explicitly decomposing the Q-value function into two functions:

\begin{itemize}
    \item State-Value function $\bm{V(s)}$ measures how good is for the agent to be in state $\bm{s}$.
    \item Advantage-Value function $\bm{A(s, a)}$ captures how good is an action compared to other actions at a given state.
\end{itemize}
 
The two functions shown in Figure \ref{fig:Dueling-DQN} \cite{Wang2016}, are combined via a special aggregation layer to produce an estimate of the state-action value function \cite{Wang2016}. The value of this function is equal to the summation of the two values produced by the two functions:

\begin{equation}
    \label{eq:dueling-dqn}
    Q(s,a) = V(s) + \big( A(s,a)  -\frac{1}{|\mathcal{A}|} \sum_{a^\prime} A(s,a) \big)
\end{equation}

The subtracted term $\bm{\frac{1}{|\mathcal{A}|} \sum_{a^\prime} A(s, a)}$ represents the mean, where $\bm{|\mathcal{A}|}$ represents the size of the vector $\bm{A}$. This term helps with identifiability, and it does not change the relative rank of the A (and hence Q) values. Additionally, it increases the stability of the optimization as the advantage function only needs to change as fast as the mean \cite{Wang2016}.

Double Dueling DQN (DD-DQN) is another DQN algorithm: it combines Dueling DQN with Double DQN to find the optimal Q-value as suggested originally by Wang \textit{et al.} \cite{Wang2016} where the output from the Dueling DQN is passed to Double DQN.

\vspace{3 pt}
\paragraph{Deep Recurrent Q-Networks (DRQN)}
\hfill \break
Deep Recurrent Q-Network (DRQN) \cite{hausknecht2015deep} is an extension of the DQN algorithm, replacing the first fully connected layer with a recurrent LSTM layer of the same size. Adding the LSTM layer requires changing the input size from a single state of the environment to multiple states (frames) as a single input, which helps to integrate information through time \cite{hausknecht2015deep}.
\vspace{6 pt}
\subsubsection{\textbf{Deep SARSA}}
\hfill \break
Basic SARSA uses Q learning and is suitable for limited states and discrete actions environments, as described in subsection \ref{LimitedStatesDiscreteActions}. On the other hand, Deep SARSA for unlimited states uses a deep neural network similar to DQN: the main difference is that SARSA computes $\bm{Q(s^\prime,a^\prime)}$ by performing the next action $\bm{a^\prime}$, which is required to calculate the current state-action $\bm{Q(s,a)}$. As shown in Figure \ref{fig:RL-Algorthims}, extensions of Deep SARSA are the same as extensions of DQN with the main difference on how to calculate the next action-state value $\bm{Q(s^\prime,a^\prime)}$.

\subsection{Environments with Unlimited States and Continuous Actions}

Although discrete actions are sufficient to move a car or UAV in a virtual environment, such actions do not provide a realistic object movement in real-life scenarios. Continuous actions describe the quantity of movement in different directions where the agent does not choose from a list of predefined actions. For example, a realistic UAV movement specifies the quantity of required change in roll, pitch, yaw, and throttle values to navigate the environment while avoiding obstacles, rather than moving UAV using one step in predefined directions: up, down, left, right, and forward.

Continuous action space requires the agent to learn a parameterized policy $\bm{\pi_\theta}$ to maximize the expected summation of the discounted rewards because it is impossible to calculate action-value for all continuous actions at different states. The problem is a maximization problem and can be solved using gradient descent algorithms to find the optimal $\bm{\theta}$. The value of $\bm{\theta}$ is updated as follows:

\begin{equation}
    \label{eq:gradient-ascent}
    \theta_{t+1} = \theta_{t} + \alpha\nabla J(\theta_{t})
\end{equation}

\noindent where $\bm{\alpha}$ is the learning rate and $\bm{\nabla}$ is the gradient.

The reward function $\bm{J}$ objective is to maximize the expected reward using a parameterized policy $\pi_\theta$ as follows \cite{sutton2018reinforcement}:

\begin{equation}
    \label{eq:policy-gradient}
    \begin{split}
        J(\pi_\theta) &= \sum_{s \in S} \rho_{\pi_\theta}(s) \; V^{\pi_\theta}(s) \\
                    &= \sum_{s \in S} \rho_{\pi_\theta}(s) \; \sum_{a\in A} Q^{\pi_\theta}(s, a) \ \pi_{\theta}(a|s) 
    \end{split}
\end{equation}

Here $\bm{\rho_{\pi_\theta}(s)}$ defines the stationary probability of $\bm{\pi_\theta}$ starting from state $\bm{s_0}$ and transitioning to future states following the policy $\bm{\pi_\theta}$ for $\bm{t}$ time steps. Finding the optimal $\bm{\theta}$ that maximizes the function $\bm{J(\pi_\theta)}$ requires finding the gradient $\bm{\nabla_{\theta} J(\theta)}$:

\begin{equation}
    \label{eq:policy-gradient-theorem}
    \begin{split}
        \nabla_{\theta} J(\theta) &= \nabla_{\theta} \biggl( \sum_{s \in S} \rho_{\pi_\theta}(s) \; \sum_{a\in A}  Q^{\pi_\theta}(s, a) \ \pi_{\theta}(a|s) \biggr) \\
                                  &\propto \sum_{s \in S} \mu(s) \; \sum_{a\in A} Q^{\pi_\theta}(s, a) \  \nabla \pi_{\theta}(a|s)\
    \end{split}
\end{equation}

Equation \ref{eq:policy-gradient-theorem} can be further rewritten in continuous episodes since $\bm{\sum_{s \in  S} \eta(s) = 1}$ as:

\begin{equation}
    \label{eq:policy-gradient-theorem2}
    \nabla_{\theta} J(\theta) = \mathbb{E}_{s \sim \rho^{\pi_\theta} , a \sim \pi_\theta} \Big [ Q^{\pi_\theta}(s, a) \; \nabla_{\theta} \ln \pi_\theta (a_t|s_t) \Big ]
\end{equation}

When the training sample is collected according to the target policy $\bm{s \sim \rho^{\pi_\theta}}$ and the expected return is generated for the same policy $\bm{\pi_\theta}$, the algorithm is referred to as \textit{on-policy algorithm}. On the other hand, in \textit{off-policy algorithms}, the training sample follows a behavior policy $\bm{\beta(a|s)}$, which is different than the target policy $\bm{\pi_\theta(a|s)}$ \cite{silver2014deterministic}, while the expected reward is generated using the target policy $\bm{\pi_\theta}$. Off-policy algorithms do not require full rejectories (episodes) for the training sample and they can reuse past trajectories. Equation \ref{eq:off-policy-gradient-theorem} \cite{silver2014deterministic} shows how the policy is adjusted to the ratio between the target policy $\bm{\pi_\theta(a|s)}$ and behaviour policy $\bm{\beta(a|s)}$. 
\begin{equation}
    \label{eq:off-policy-gradient-theorem}
    \nabla_{\theta} J(\theta) = \mathbb{E}_{s \sim \rho^{\beta} , a \sim \beta} \Big [ \frac{\pi_\theta(a|s)}{\beta_{\theta}(a|s)} Q^{\pi_\theta}(s, a) \; \nabla_{\theta} \ln \pi_\theta (a_t|s_t) \Big ]
\end{equation}

The policy gradient theorem shown in equation \ref{eq:policy-gradient} \cite{sutton2000policy} considered the fundamental base of distinct Policy Gradients (PG) algorithms such as REINFORCE \cite{williams1992simple}, Actor-Critic algorithms \cite{konda2000actor}, Trust Region Policy Optimization (TRPO) \cite{schulman2015trust}, and Phasic Policy Gradient \cite{raileanu2021decoupling}, Stein Variational Policy Gradient \cite{liu2017stein}, Proximal Policy Optimization (PPO) \cite{schulman2017proximal}, and many others.
\vspace{6 pt}
\subsubsection{\textbf{REINFORCE}}
\hfill \break
REINFORCE is an acronym for \textbf{RE}ward \textbf{I}ncrement $=$ \textbf{N}onnegative \textbf{F}actor $\times$ \textbf{O}ffset \textbf{R}einforcement $\times$ \textbf{C}haracteristic \textbf{E}ligibility \cite{williams1992simple}. REINFORCE is a Monte-Carlo policy gradient algorithm that works with the episodic case. It requires a complete episode to obtain a sample proportional to the gradient, and updates the policy parameter $\bm{\theta}$ with the step size $\bm{\alpha}$. Because $\bm{\mathbb{E}_{\pi}[G_t|S_t, A_t] =  Q^{\pi}(s, a)} $, REINFORCE can be defined as \cite{sutton2018reinforcement}:

\begin{equation}
    \label{eq:REINFORCE}
    \nabla_{\theta} J(\theta) = \mathbb{E}_{\pi} \Big [G_t \; \nabla_{\theta} \ln \pi_\theta (A_t|S_t) \Big ]
\end{equation}

REINFORCE uses the Monte Carlo method, which suffers from high variance and, consequently, has slow learning \cite{williams1992simple}. Adding a baseline to REINFORCE reduces the variance and speeds up learning while keeping the bias unchanged by subtracting the baseline value from the expected return $\bm{G_t}$ \cite{sutton2018reinforcement}.
\vspace{6 pt}
\subsubsection{\textbf{Trust Region Policy Optimization (TPRO)}}
\hfill \break
Trust Region Policy Optimization (TRPO) \cite{schulman2015trust} is a PG algorithm that improves the performance of gradient descent by taking more extensive steps within trust regions defined by a constraint of KL-Divergence and performs the policy update after each trajectory rather than after each state. Proximal Policy Optimization (PPO) \cite{schulman2017proximal} can be considered an extension of TRPO; it imposes the constraint as a penalty and clips the objective to ensure that the optimization is carried out within the predefined range \cite{shin2019obstacle}.

Phasic Policy Gradient (PPG) \cite{cobbe2020phasic} extends PPO by including a periodic auxiliary phase which distills features from the value function into the policy network to improve training. This auxiliary phase enables feature sharing between the policy and value function while decoupling their training.

\vspace{6 pt}
\subsubsection{\textbf{Stein Variational Policy Gradient (SVPG)}}
\hfill \break
Stein Variational Policy Gradient (SVPG) \cite{liu2017stein} applies the Stein variational gradient descent (SVGD) \cite{liu2016stein} to update the policy parameterized by $\bm{\theta}$, which reduce variance and improves convergence. SVPG improves the average return and data efficiency when used on top of REINFORCE and advantage actor-critic algorithms \cite{liu2017stein}.
\vspace{6 pt}
\subsubsection{\textbf{Actor-Critic}}
\hfill \break
Actor-Critic algorithms are a set of algorithms based on policy gradients theorem that consist of two components: 
\begin{enumerate}
    \item An Actor responsible for adjusting the parameter $\bm{\theta}$ of the policy $\bm{\pi_\theta}$
    \item A Critic which employs a parameterized vector $\bm{w}$ to estimate the value-function $\bm{Q^{w}(s_t,a_t) \approx Q^{\pi}(s_t,a_t)}$ using a policy evaluation algorithm such as temporal-difference learning \cite{silver2014deterministic}
\end{enumerate}

The actor can be described as the network trying to find the probability of all available actions and select the action with the highest value, while the critic can be described as a network evaluating the selected action by estimating the value of the new state resulted from performing the action. Different algorithms fall under the actor-critic category; the main ones are described in the following subsections.
\vspace{3 pt}
\paragraph{Deterministic Policy Gradients (DPG) Algorithms}
\hfill \break
All deterministic policy gradients algorithms model the policy as a deterministic policy $\bm{\mu(s)}$, rather than stochastic policy $\bm{\pi(s,a)}$ that is modeled over the action's probability distribution. We described earlier in Equation \ref{eq:policy-gradient}, the objective function under a selected policy $\bm{J(\pi_\theta)}$ to be $\bm{\sum_{s \in S} \rho_{\pi_\theta}(s) \; V^{\pi_\theta}(s)}$; however, a deterministic policy is a special case of stochastic policy, where the objective function of the target policy is averaged over the state distribution of the behaviour policy as described in equation \ref{eq:deterministic-policy-gradient} \cite{silver2014deterministic}.

\begin{equation}
    \label{eq:deterministic-policy-gradient}
    \begin{split}
        J_{\beta}(\mu_{\theta}) &= \int_{S} \rho^{\beta}(s) \ V^{\mu}(s) \ ds \\
                                &= \int_{S} \rho^{\beta}(s) \ Q^{\mu}(s,\mu_{\theta}(s)) \ ds
    \end{split}
\end{equation}

In the off-policy approach with a stochastic policy, importance sampling is often used to correct the mismatch between behaviour and target policies. However, because the deterministic policy gradient removes the integral over actions, we can avoid importance sampling and the gradient becomes:

\begin{equation}
    \label{eq:deterministic-policy-gradient-theorem}
    \begin{split}
        \nabla_{\theta} J_{\beta}(\mu_{\theta}) &\approx \int_{S} \rho^{\beta}(s) \ \nabla_{\theta} \ \mu_{\theta}(a|s) \ Q^{\mu}(s,\mu_{\theta}(s)) \ ds \\
                                                &= \mathbb{E}_{s \sim \rho^{\beta}} \ \Big [ \nabla_{\theta} \ \mu_{\theta}(s) \nabla_{a} Q^{\mu}(s,a)|_{a=\mu_{\theta}(s)} \Big ]
    \end{split}
\end{equation}

Different algorithms build on DPG with improvements; for example, Deep Deterministic Policy Gradient (DDPG) \cite{lillicrap2015continuous} adapts DQN to work with continuous action space and combines it with DPG. On the other hand, Distributed Distributional DDPG (D4PG) \cite{barth2018distributed} adopts distributed settings for DDPG with additional improvements such as using N-step returns and prioritized experience replay \cite{barth2018distributed}. Multi-agent DDPG (MADDPG) \cite{lowe2017multi} is another algorithm that extends DDPG to work with multi-agents, where it considers action policies of other agents and learns policies that require multi-agent coordination \cite{lowe2017multi}.

Twin Delayed Deep Deterministic (TD3) \cite{fujimoto2018addressing} builds on Double DQN and applies to DDPG to prevent the overestimation of the value function by taking the minimum value between a pair of critics \cite{fujimoto2018addressing}.

\vspace{3 pt}
\paragraph{Advantage Actor-Critic (A3C)}
\hfill \break
Asynchronous Advantage Actor-Critic (A3C) \cite{mnih2016asynchronous} is a policy gradient algorithm that uses multi-threads, also known as agents or workers, for parallel training. Each agent maintains a local policy $\bm{\pi_\theta(a_t|s_t)}$ and an estimate of the value function $\bm{V_\theta(s_t)}$. The agent synchronizes its parameters with the global network having the same structure. 

The agents work asynchronously, where the value of the network parameters flows in both directions between the agents and the global network. The policy and the value function are updated after $t_{max}$ actions or when a final state is reached \cite{mnih2016asynchronous}.

Advantage Actor-Critic (A2C) \cite{mnih2016asynchronous} is another policy gradient algorithm similar to A3C, except it has a coordinator responsible for synchronizing all agents. The coordinator waits for all agents to finish their work either by reaching a final state or by performing $\bm{t_{max}}$ actions before it updates the policy and the value function in both direction between the agents and the global network.


Actor-Critic with Experience Replay (ACER) is an off-policy actor-critic algorithm with experience replay that uses a single deep neural network to estimate the policy $\pi_\theta(a_t|s_t)$ and the value function $V_{\theta_v}^{\pi}(s_t)$ \cite{wang2016sample}. The three main advantages of ACER over A3C are \cite{mnih2016asynchronous}: 1) it improves the truncated importance sampling with the bias correction, 2) it uses stochastic dueling network architectures, and 3) it applies a new \textit{trust region policy optimization} method \cite{wang2016sample}.

ACER uses an improved Retrace algorithm as described in Equation \ref{eq:retrace-algorithm} \cite{munos2016safe} by applying truncated importance sampling with bias correction technique and using the value $\bm{Q^{ret}}$ as the target value to train the critic by minimizing the L2 error term \cite{wang2016sample}. In ACER, the gradient $\bm{\hat{g}_{t}^{acer}}$ is computed by truncating the importance weights by a constant $\bm{c}$, and subtracting $\bm{V_{\theta_v}(s_t)}$ to reduce variance: this is denoted in Equation \ref{eq:acer-gradiant} \cite{wang2016sample}.
\begin{equation}
    \label{eq:retrace-algorithm}
    \begin{split}
    Q^{ret}(s_t,a_t) &= r_t + \gamma \bar{\rho}_{t+1} \big [ Q^{ret}(s_{t+1},a_{t+1}) - Q(s_{t+1},a_{t+1})    \big ] \\
    & \;\; + \gamma V(s_{t+1})
    \end{split}
\end{equation}
\begin{equation}
    \label{eq:acer-gradiant}
    \begin{split}
    \hat{g}_{t}^{acer} & = \bar{\rho}_t \nabla_{\theta} \ln \pi_\theta(a_t|s_t) \big [Q^{ret}(s_t , a_t) - V_{\theta_v}(s_t) \big ] \\
    & \;\; + \underset{a \sim \pi}{\mathbb{E}} \Big (  \big [ \frac{\rho_t(a) - c}{\rho_t(a)}  \big ] \nabla_{\theta} \ln \pi_\theta(a_t|s_t) \\
    & \;\;\;\;\;\; \big [ Q_{\theta_v}(s,_t,a_t) - V_{\theta_v}(s_t) \big ] \Big )
    \end{split}
\end{equation}

Actor-Critic using Kronecker-Factored Trust Region (ACKTR) \cite{wu2017scalable} is another extension of A3C \cite{mnih2016asynchronous}, which optimizes both the actor and critic by using Kronecker-factored approximation curvature (K-FAC) \cite{martens2015optimizing}. It provides an improved computation of the natural gradients by allowing the covariance matrix of the gradient to be efficiently inverted \cite{wu2017scalable}.

\vspace{3 pt}
\paragraph{Soft Actor-Critic (SAC)}
\hfill \break
Soft Actor-Critic (SAC) aims to maximize the expected reward while maximizing the entropy \cite{haarnoja2018soft}. SAC ameliorates the maximum expected sum of rewards defined through accumulating the reward over states transitions $J(\pi) = \sum_{t=1}^{T} \mathbb{E}_{s \sim \rho^{\pi} , a \sim \pi} \Big [ r(s_t,a_t) \Big ]$ by adding the expected entropy of the policy over $\rho_\pi(s_t)$ \cite{haarnoja2018soft}. Equation \ref{eq:sac-entropy} shows a generalized entropy objective, where the temperature parameter $\alpha$ controls the stochasticity of the optimal policy through defining the relevance of the entropy $\mathcal{H}(\pi(.|s_t))$ term to the reward \cite{haarnoja2018soft}. 
\begin{equation}
    \label{eq:sac-entropy}
    J(\pi) = \sum_{t=1}^{T} \mathbb{E}_{s \sim \rho^{\pi} , a \sim \pi} \Big [ r(s_t,a_t) + \alpha \mathcal{H}(\pi(.|s_t)) \Big ]
\end{equation}

SAC uses two separate neural networks for the actor and critic, and applies function approximators to estimate a soft Q-function $\bm{Q_\theta(s_t,a_t)}$ parameterized by $\bm{\theta}$, a state value function $\bm{V_\psi(s_t)}$ parameterized by $\bm{\psi}$, and an adjustable policy $\bm{\pi_\phi(a_t|s_t)}$ parameterized by $\bm{\phi}$.

\vspace{3 pt}
\paragraph{Importance Weighted Actor-Learner Architecture (IMPALA)}
\hfill \break
Importance Weighted Actor-Learner Architecture (IMPALA) \cite{espeholt2018impala} is an off-policy learning algorithm that decouples acting from learning and can be used in two different setups: 1) single learner and 2) multiple synchronous learners. 

Using a single learner and multiple actor setup, each actor generates trajectories and sends each trajectory to the learner, and receives the updated policy before starting a new trajectory. The learner learns from the actors simultaneously by saving the received trajectories from the actors in a queue and generating the updated policy. Nevertheless, actors might learn an older model because actors are not aware of each other and because of the lag between the actors and the learner. To resolve this issue, IMPALA uses a novel v-trace correction method that considers a truncated importance sampling (IS), which is the ratio between the learner policy $\bm{\pi}$ and the actor current policy $\bm{\mu}$.
Similarly, in multiple synchronous learners, the policy parameters are distributed across multiple learners that work synchronously through a master learner \cite{espeholt2018impala}.

\section{Conclusion} \label{sec:conclusion}


Deep Reinforcement Learning has shown advancement in solving sophisticated problems in real-life scenarios. The environment type of the application has a vital role in selecting an appropriate RL algorithm that provides good results and performance. In this work, we have identified three environment types based on the number of actions and states: 1) Limited states and discrete actions, 2) Unlimited states and discrete actions, and 3) Unlimited states and continuous actions.

Environments with a limited number of states and limited actions are considered austere environments and can be solved using Q-learning and SARSA. Complex environments have unlimited states representing the environment, and applying the appropriate algorithm depends on the number of actions. If the actions are limited (discrete), the value-based algorithms such as DQN and its variations would be the choice. However, if the actions are continuous, the policy gradient algorithms are appropriate as they can learn a parameterized policy that approximates the solution. This classification helps researchers and practitioners select appropriate RL algorithms for their studies and applications.


Further investigation of algorithms performance in different use case scenarios is needed: the algorithms should be compared in respect to accuracy, convergence, computational resources, and ease of use. Moreover, diverse use cases and requirements should be considered in the evaluation.   

\bibliographystyle{IEEEtran}
\bibliography{references.bib}

\end{document}